\documentclass{article}

\usepackage[nonatbib,final]{neurips_2023}

\usepackage[utf8]{inputenc} 
\usepackage[T1]{fontenc}    
\usepackage{hyperref}       
\usepackage{url}            
\usepackage{booktabs}       
\usepackage{multirow}
\usepackage{amsfonts}       
\usepackage{nicefrac}       
\usepackage{microtype}      
\usepackage{xcolor}         
\usepackage[textsize=tiny]{todonotes}
\usepackage{algpseudocode}
\usepackage{url}

\title{Cooperative AI via Decentralized Commitment Devices}

\author{
  Xinyuan Sun \\
  Flashbots \\
  \texttt{Xinyuan@flashbots.net} \\
  \And
  Davide Crapis \\
  Robust Incentives Group \\
  Ethereum Foundation \\
  \texttt{davide@ethereum.org} \\
  \And
  Matt Stephenson \\
  Pantera Capital \\
  \texttt{matt.stephenson@panteracapital.com} \\
  \And
  Barnab\'e Monnot \\
  Ethereum Foundation \\
  \texttt{barnabe.monnot@ethereum.org} \\
  \And
  Thomas Thiery \\
  Ethereum Foundation \\
  \texttt{thomas.thiery@ethereum.org} \\
  \And
  Jonathan Passerat-Palmbach \\
  Flashbots \\
  Imperial College London \\
  \texttt{jonathan@flashbots.net} \\
}

\begin{document}

\maketitle

\begin{abstract}
Credible commitment devices have been a popular approach for robust multi-agent coordination. However, existing commitment mechanisms face limitations like privacy, integrity, and susceptibility to mediator or user strategic behavior. It is unclear if the cooperative AI techniques we study are robust to real-world incentives and attack vectors. However, decentralized commitment devices that utilize cryptography have been deployed in the wild, and numerous studies have shown their ability to coordinate algorithmic agents facing adversarial opponents with significant economic incentives, currently in the order of several million to billions of dollars. In this paper, we use examples in the decentralization and, in particular, Maximal Extractable Value (MEV) ~\cite{daian2020flash} literature to illustrate the potential security issues in cooperative AI. We call for expanded research into decentralized commitments to advance cooperative AI capabilities for secure coordination in open environments and empirical testing frameworks to evaluate multi-agent coordination ability given real-world commitment constraints.
\end{abstract}

\section{Introduction}

AI coordination and safety are a necessity~\cite{russell2019human}~\cite{amodei2016concrete}~\cite{dafoe2020open}. With an unrestricted empowerment of algorithmic agents, the lack of coordinated interactions between them can give rise to considerable negative externalities. For example, collusive pricing in e-commerce marketplaces or gas stations~\cite{assad2020algorithmic} can hurt consumer welfare~\cite{ecommercedoj}~\cite{chen2016empirical}~\cite{wieting2021algorithms}, emergent equilibria~\cite{edelman2007strategic} from auto-bidders in ad auctions could lead to revenue loss~\cite{kolumbus2022auctions}~\cite{mehta2023auctions}~\cite{banchio2022artificial}, high-frequency trading bots can seriously reduce the health of markets~\cite{budish2015high}~\cite{easley2011microstructure}, and sniper bots on online concert ticket platforms poses profound user experience concerns~\cite{ticketmasters}~\cite{ticketmastersbill}.

However, most existing approaches to both coordination and safety problems presume critical use of traditional social technologies like policies and regulation~\cite{hacker2023regulating}~\cite{regulation}~\cite{edwards2022regulating}, which are insufficient and too slow for coordinating games with a mixture of human and algorithmic participants, especially in zero-shot scenarios~\cite{hu2020other}. For example, regulations over financial transactions~\cite{mlexecution} and ad auctions~\cite{lawsuitgoogle} have traditionally failed in coordinating or aligning the incentives of algorithmic agents. A potential way to foster secure and reliable multi-agent cooperation, filling the gaps left by traditional systems, lies in credible commitment devices~\cite{kalai2010commitment}. Existing works on cooperative AI~\cite{conitzer2022foundations} have been focusing on enabling coordination between AIs when agents can credibly commit to running certain source codes~\cite{oesterheld2022similarity}~\cite{critch2019parametric} and allow simulation based on those committed source codes~\cite{kovarik2023game}~\cite{tennenholtz2004program}~\cite{critch2022cooperative}~\cite{digiovanni2023commitment}\footnote{Here, commitment means an enforcement whose knowledge can change incentives, implementing all feasible and individually rational strategies in a game~\cite{kalai2010commitment}.}. Many multi-agent reinforcement learning (MARL) works also implicitly assume that there is a credibly committed training algorithm of agents that other agents may have access to~\cite{foerster2017learning} (or even control over~\cite{claus1998dynamics}) to achieve zero-shot coordination~\cite{treutlein2021new}~\cite{hu2020other}. Finally, studies have shown that agents learn the ability to robustly coordinate via contractual commitments across a large class of coordination games~\cite{christoffersen2022get}~\cite{hughes2020learning}~\cite{mcaleer2021improving}.

One popular way to implement credible commitments is via decentralized trust by leveraging cryptographic schemes ~\cite{shi2022can} (such as multiparty computation~\cite{Lindell_2020}, zero-knowledge proofs~\cite{Bitansky_Canetti_Chiesa_Goldwasser_Lin_Rubinstein_Tromer_2014}, etc.) to give agents commitment power~\cite{virgil}. For example, one could use basic cryptographic signatures and commitment schemes~\cite{ferreira2020credible} to circumvent strong negative results~\cite{akbarpour2020credible} in auction mechanisms where the auctioneer is a rational agent. Moreover, combining cryptography with distributed systems allows one to implement smart contracts~\cite{buterin2014next} and use them as public broadcast channels~\cite{choudhuri2017fairness} that enable a large set of credible multi-agent mediations to be implemented~\cite{chitra2023credible}~\cite{matsushima2020mechanism}. There has been a rich literature~\cite{daian2020flash}~\cite{mazorra2023cost}~\cite{ferreira2022credible}~\cite{landis2023side}~\cite{landis2023stackelberg}~\cite{kulkarni2022towards}~\cite{babel2023lanturn} on the ability and security of those credible commitment devices deployed in the real world, especially when they face large-scale strategic behavior by a mixture of both human and algorithmic agents with \textit{billion dollar} amount incentives~\cite{MEV-ExploreV1}~\cite{qin2021attacking}~\cite{MEV-Dashboard}.

In this context, the study of decentralized commitment devices provides a pragmatic framework for understanding the potential limitations of the commitment assumption in multi-agent coordination settings. Namely, \textit{when we implement and deploy AI agents in the real world to engage in few-shot coordination tasks, what new behaviors should we expect to emerge?  Are the assumptions we make in MARL and cooperative AI realistic? What would the security properties of those commitment devices be? And with those security properties of the commitment devices in mind, will AIs still coordinate robustly?} 

In this paper, we first answer those questions via a few examples that illustrate the synergy between the study of decentralized commitment devices and multi-agent security. The parallels we draw should be a stepping stone for future research agendas in evaluating robust zero-shot coordination games with AI agents. We end by calling for action on evaluating multi-agent coordination ability in the presence of a decentralized commitment device, as we believe that, aside from theory, the "crypto-economic commitments sandbox" also emerges as a \textit{tangible platform to test multi-agent coordination with real-world incentives}.


\section{Congestion Games}\label{sec:congestion_games}

Many games studied in the decentralized commitment device literature pose interesting questions about the security properties of multi-agent coordination. One such example is congestion games, a major area of interest for computational game theory and multi-agent systems. 

Congestion games are a canonical example of a mixed-interest game with competitive and cooperative motives at play. Recent studies in artificial intelligence have focused on algorithms for learning equilibria and welfare effects~\cite{kolumbus2022and}~\cite{lowe2017multi}~\cite{lu2022model}~\cite{kim2022game} in different subclasses of games with different assumptions on the information available to agents.  Congestion games arise naturally in various contexts where sophisticated agents/bots interact. The most notable ones beyond robotics are online platforms (from online advertising~\cite{kolumbus2022and} to Decentralized Finance (DeFi)~\cite{werner2022sok} on blockchains). ~\cite{kulkarni2022towards} models asset exchange on decentralized commitment devices as a routing game.

In the DeFi context, a rich evolving topology has emerged, with new nodes (assets) and new routes (liquidity pools~\cite{angeris2021analysis}, exchanges,  bridges~\cite{belchior2021survey}) being added and a differential set of more and less sophisticated/informed agents that create and route orders (see Figure \ref{fig-routing} in the Appendix). Specifically, traders delegate their decision-making to a program (called a "transaction") that decides on the creation and routing of orders. Complex behaviors have emerged, such as sophisticated strategies~\cite{daian2020flash} for \textit{frontrunning} or \textit{backrunning} orders routed suboptimally~\cite{angeris2022optimal} from an ecosystem of self-interested agents and organizations.

Strategies like those mentioned above are real-world examples of Maximal Extractable Value (MEV) ~\cite{daian2020flash}, which refers to the externalities caused by mediators of commitment devices playing strategically. In this case, the mediator is the miner~\cite{buterin2014next} or the block producer~\cite{bahrani2023transaction} responsible for ordering user transactions (commitments) and including them in the canonical chain (making them credible). The mediator can use its informational advantage from seeing other agents' strategies and parameters by simulating outcomes to identify an optimal action to insert right before (frontrunning) or right after (backrunning) to extract economic value from other agents.

Here, the multi-agent security problem comes from the fact that some agents (mediators) can simulate other agents an infinite number of times, which encourages cooperation~\cite{oesterheld2022similarity} but also creates an asymmetry in payoffs. Other examples of MEV include an auctioneer in a second-price auction inserting a shill bid after observing all bidders' bids to extract value from the winner ~\cite{vickrey1961counterspeculation}~\cite{akbarpour2020credible}.


In the following sections, we explore the limits of real-world commitments in both AI and blockchain. We highlight the disparities between prevailing notions of commitments and the requisites of real-world commitments and emphasize the need for robust solutions to address these limitations.

\section{Limits of Real-world Commitments}\label{limits-of-real-world-commitments}


When considering cooperative AI agents, there is often an implicit assumption of the existence of a trusted
centralized mediator\footnote{Here, the mediators should be explicitly interpreted as in mediated equilibrium~\cite{monderer2009strong} where it has commitment power to implement (coarse) correlated strategies.} responsible for orchestrating interactions among
agents \cite{tennenholtz2004program}~\cite{foerster2017learning}. This central mediator assumes a critical role, but recent antitrust litigation against platforms like Google's ad auction~\cite{lawsuitgoogle} serves as a reminder of the risks associated with entrusting a single all-powerful centralized entity with coordinating actions among agents despite the mediator has a long-term reputation at stake. Even when agents don't explicitly use commitments (see elaboration in Appendix ~\ref{appendix} on a concrete game) or they take turns acting as mediators during contract negotiations, the risk of collusion persists. In reality, relying on identity verification to mitigate this issue is impractical, as agents and bots are more susceptible to Sybil attacks than human agents. 


In real-world settings, not
all applications or agents will tolerate simulation or parameter sharing (for a concrete exploit on this involving shared world models, see Appendix ~\ref{appendix}).
This constraint suggests we should consider the use of Privacy Enhancing Technologies (PETs) \cite{PPTTT_2019}
to facilitate contract formation. Moreover, when formulating assumptions regarding
cooperative agents, it is imperative to consider the constraints
introduced by PETs, including their impact on performance, hardware
requirements, and the expressiveness of contracts.




Both AI and MEV settings share common limitations. Credible commitment
devices, whether permissionless or not, must offer real-world validity
guarantees. 
To instill trust in such mechanisms, users need assurance on how commitments maintain integrity once deployed. Two approaches exist – constructing commitment
devices with inherent integrity obtained via verifiable cryptographic tools such as SNARKs (Succinct 
Non-interactive ARguments of Knowledge) \cite{Bitansky_Canetti_Chiesa_Goldwasser_Lin_Rubinstein_Tromer_2014}, or optimistic reliance on monitoring and penalizing defection~\cite{ostrom1990governing}.

Presently, real-world deployments predominantly lean toward the
optimistic approach due to technical constraints. Constructing integrity guarantees by design introduces complexity and higher overhead, existing methods can now support deep neural networks of the size of MobileNet\cite{sandler2018mobilenetv2} or GPT2\cite{gpt2} with a 10-20X inference time overhead on CPU \cite{Kang_Hashimoto_Stoica_Sun_2022}, which leaves a lot of room for further software and hardware speed-ups. 
While computationally tractable, optimistic approaches fall short in scenarios with fat-tail payoffs. There have been incidents where \$20 million loss was caused by exploiting payoff asymmetry in commitments~\cite{mevboostattack}.







\section{Security of Commitments in Cooperative AI}

We underscore some security complications of commitment devices mentioned above, evident across a broad spectrum of existing cooperative AI work. 

First, in situations where cooperation relies on hard-wired heuristics in the training process or the learning algorithm, such as augmenting intrinsic motivation~\cite{jaques2019social} by changing reward functions or employing Centralized Training with Decentralized Execution (CTDE)\cite{oliehoek2016concise}\cite{lowe2017multi}, the necessity of a credible commitments to opponents' training process prior to an agent's interaction is paramount. This necessity vests significant trust in the mediator providing this proof, thereby presenting a potential vulnerability, as mediators can effectively dictate the equilibrium the agents converge to in general-sum coordination games (see Appendix ~\ref{sec:methods_commitment} for an illustrative example). This incentivizes collusion between the mediator and agents to manipulate the training process or provide non-factual proofs.

Agents using algorithms such as LOLA~\cite{foerster2017learning} or COLA~\cite{willi2022cola}, which assumes access to opponents' learning parameters, would similarly require a verifiable proof that its opponents are indeed running the reported parameters. This proof can be viewed as a commitment by either the opponent or the mediator, and it suffers from credibility problems. In reality, agents may not want to fully expose their parameters due to privacy or commercial reasons, leading them to entrust proof delegation to a mediator. However, this incentivizes the mediator to accept bribes from agents to misreport parameters as it can earn this profit risk-freely since no single agent can provide evidence of cheating~\cite{akbarpour2020credible}. Of course, those algorithms can evade the commitment credibility problem via opponent modeling using techniques like behavior cloning~\cite{torabi2018behavioral} at the cost of performance, posing an interesting meta-coordination problem: since parameter sharing is incompatible with privacy, agents will likely resort to using entrusted mediators, and because of strategic mediator behavior, agents may resort back to using opponent modeling, giving up potential coordination gains.

In credible commitments, the problem of Maximal Extractable Value (MEV)~\cite{daian2020flash} is even more apparent when agents explicitly use correlation devices to coordinate. For example, if cooperative AI approaches where agents learn payment contracts~\cite{christoffersen2022get} or have access to signals from communication channels~\cite{cigler2013decentralized} are implemented in reality, we would likely see agents engage in MEV behavior. For example, in mediated multi-agent reinforcement learning~\cite{ivanov2023mediated}, the mediator is explicitly modeled as a learning agent who can make credible commitments and takes reports from agents about their observation/reward/action pairs to design an optimal contract. It is not uncommon for users to play the meta-game here and strategically misreport inputs to the mediator or  their own agents~\cite{kolumbus2022and}~\cite{kolumbus2022auctions}. 

Moreover, it has been demonstrated in existing MARL works that agents that coordinate using correlation devices could converge to unfair outcomes where one agent learns to propose a contract that extracts close to all of the surplus from coordination and distributes only $\epsilon$ gain the rest~\cite{christoffersen2022get}. This is a longstanding problem in game theory where the mechanism designer, or the Stackelberg player, gets to extract the entire surplus. If those cooperative AI systems are actually deployed, we expect to see agents explicitly optimizing for being the ``Stackelberg player''. For example, they could disrupt the communication network via a Denial of Service (DoS) attack with their commitment such that the mediator does not see other agents' commitments or chooses to send manipulative messages~\cite{blumenkamp2021emergence}. 

Alternatively, instead of directly manipulating the mediator/correlation device, the agents may learn to do long-term optimization by learning to impact which equilibrium the policies converge to, with the resulting equilibrium formalized as \textit{active equilibrium}~\cite{kim2022influencing}~\cite{kim2022game}. It has also been demonstrated that agents can learn to play the meta-game of impacting equilibrium selection in learning convergence rather than myopically optimizing for reward~\cite{xie2021learning}~\cite{lu2022model}. Here, the problem is not with the credibility of the commitment but rather whether the inputs to the commitment reflect the true state of the world (See Appendix ~\ref{appendix} for a concrete example). Such problems are exacerbated in the face of imperfectly monitored censorship and insertion~\cite{akbarpour2020credible}.

Another area of MARL research has specifically focused on zero-shot coordination settings \cite{hu2021otherplay} in which agents do not explicitly rely on a central controller or commitment devices to learn and coordinate their actions. Instead, the structure of a decentralized partially observable Markov Decision Processes (Dec-POMDPs, \cite{tamingdecPOMDPs}) itself needs to be used for coordination. This process is controlled by multiple distributed agents, each with possibly different information, which prohibits them from coordinating based on state, action, and observation labels. A variety of problems can be framed as decentralized control of a Markov process, ranging from multiple robots playing soccer \cite{decPOMDPsoccer} to collision-free navigation of self-driving cars \cite{pomdpselfdriving} , and agents playing Hanabi \cite{Bard_2020} . This research area has unveiled specific challenges tied to decentralized coordination: Repeated interactions cause agents to adopt arbitrary and often indecipherable conventions based on fragile assumptions about other agents’ actions. This can result in failures when paired with independently trained agents or humans at test time, underscoring the importance of overseeing and regulating agents' policies in real-world scenarios. Though elegant solutions like Off Belief Learning (OBL) have been presented in the MARL litterature \cite{hu2021offbelief}, we envision that credible commitment devices could be pivotal in monitoring and verifying agents’ policies, to ensure they’re not derived from spurious correlations that that might be exploited as attack vectors.

\section{Call for Action}
In this work, we have addressed the potential of studying decentralized commitment devices in revealing a plethora of security issues inherent in existing cooperative AI endeavors. We believe the study of those devices heralds a promising horizon for enhancing multi-agent systems' robustness and security foundations.

We also see decentralized commitment devices as a valuable addition to our arsenal of techniques to achieve more secure cooperative AI. Even though it has been demonstrated that decentralized commitment devices can mitigate such security issues of centralized mediators~\cite{monderer2009strong}~\cite{shi2022can}~\cite{ferreira2020credible}~\cite{chitra2023credible}, a significant research gap persists in their application to ameliorate multi-agent security problems within the domain of cooperative AI~\cite{dafoe2020open}.

We advocate for intensified exploration into the nexus between decentralized commitment devices and cooperative AI. A critical need exists for empirical demonstrations of AI’s coordination capacity across an expansive array of general-sum games utilizing decentralized commitments alongside a comprehensive evaluation of the real-world constraints inherent in these commitments. This call to action echoes the imperative to bolster AI's capability to do few-shot coordination in environments with commitments and robustly defend against security attacks on those commitments.

It is also pivotal to traverse beyond theoretical discussions, propelling the implementation of extant cooperative AI research atop a decentralized commitment device framework. This strategic move will invariably augment the exposure to both adversarial and real-world incentives, offering a more holistic and realistic testing ground for cooperative AI, transcending the confines of an insulated environment.

The fulfillment of these objectives will significantly advance our understanding and capabilities in cooperative AI, underscoring its potential to flourish securely and efficiently in real-world applications, thus contributing profoundly to the broader AI research landscape. As we advance in the field of multi-agent systems, it is essential to recognize the role that decentralized commitments can play.






\newpage
\bibliographystyle{splncs04}
\bibliography{main}
\newpage

\appendix

\section{DeFi network and routing}

\begin{figure}[hbt!]
    \centering
    \includegraphics[width=0.5\linewidth]{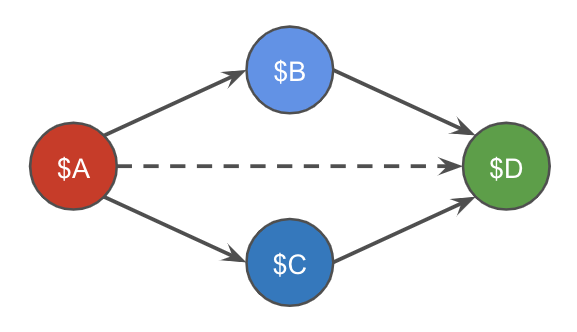}
    \caption{Example DeFi network with 4 assets and routes between them. Full lines are smart contract exchanges and dotted line is a (potential) direct swap.}
    \label{fig-routing}
\end{figure}

\section{Where does the Commitment Happen?}\label{appendix}
In MARL and Cooperative Game Theory more generally, agents are sometimes said to act as though they are committed, but the commitment lives outside the model. For example, Coarse Correlated Equilibrium \cite{moulin1978strategically} is an equilibrium refinement popular in MARL. One appealing feature is that a commonly used algorithm, external regret minimization, converges quickly to a CCE. Projected Gradient Descent in a MARL context results in a CCE as well.\cite{harris2022meta} It is a highly tractable equilibrium concept, but it has some unintuitive features related to commitment.\footnote{This was noted by the original authors who observe that their proposal is "not, strictly speaking, non-cooperative", which is to say that it leaves some strategic elements of the game unmodeled.} We will discuss it in some depth as an illustrative example of how difficult commitment can be to achieve in practice.

CCE requires a signal that is commonly observed by the agents, and it further requires that, within the game, agents cannot condition on their own behavior. This latter condition presumes strong commitment: if a player is about to choose $x$ based on the commonly observed signal, a CCE assumes that the player is not allowed to instead choose an option $y$ even if they are certain $y$ will give higher payoffs. To observe the force of this assumption, and consider plausible implementations, we will illustrate with an example.

\subsection{CCE in a Stop Light Game}
Consider the Stop Light Game (adapted from \cite{skyrms2023quasi}, with a signal set $S=\{FS,SF,SC,CS\}$, and a probability distribution function $\pi: S \to [0,1]$. If the probabilities for each signal are $ \pi(FS) =  \pi(SF) = \frac{1}{3}$ and $\pi(SC) = \pi(CS) = \frac{1}{6} $, each player obeying the signals--e.g. choosing (Fast, Stop) conditional on signal $FS$--is a CCE.

\begin{table}[h]
\centering
\begin{tabular}{|c|c|c|c|}
\hline
 & Fast & Caution & Stop \\ \hline
Fast & (0,0) & (3,1) & (7,2) \\ \hline
Caution & (1,3) & (2.1,2.1) & (6,2) \\ \hline
Stop & (2,7) & (2,6) & (4,4) \\ \hline
\end{tabular}
\caption{Payoff table for stop light game}
\label{tab:payoff_table}
\end{table}

However, note that whenever Column (Row) receives the signal $CS$ ($SC$) they strictly prefer to choose Fast instead of Caution. Doing so gives them a certain payoff increase of 1. If players could unilaterally defect to their preferred choice, such Coarse Correlated Equilibria would be anti-correlated.

Players who could bribe the mediator in some way can also improve their outcome, even without formally defecting. If Column bribes the mediator to signal $FS$, Column receives a guaranteed payoff of 7 which is higher than their CCE payoff (of 4.3 in expectation.) In principle, since the signal is ostensibly random and thus an observed $\hat \pi(FS)=1$ is not ruled out, the mediator could profit about 2 per draw.

\subsection{Methods of Commitment}\label{sec:methods_commitment}

One way to improve the incentive compatibility of a CCE in the Stop Light game would be to credibly commit to destroy some amount greater than 1 if the non-recommended strategy is chosen. This is effectively the "decentralized commitment device" method. 

Another is to try and restrict the choices of the agents in some other way. Among machines, one commitment would be to not use an algorithm that penalizes losses at each decision node, such as swap regret minimization.\footnote{Swap regret minimization is of the form ``every time I chose action i, I should have chosen
action j instead'', and thus would choose $Disobey|S=CS$ in our reduced stop light game.\cite{blum2008regret}}
But this would seem to just beg the question of why swap regret minimization is not chosen if it provides better payoffs in the actual game. 

A further possibility is conditional commitment by mutual inspection, as in Tennenholtz's Program Equilibria \cite{tennenholtz2004program}. Two cars at an intersection could potentially settle on $(Caution,Stop)|S=(CS)$  by inspecting the others' program and conditioning on it being the same as well as the signal $CS$. But a player that somehow learned to misunderstand $CS$ as, say, $FS$ does strictly better. And so too perhaps would a car owner who was able to modify their car to misinterpret $CS$ as $FS$. 

Maybe such errant behavior could be ruled out if it required a changing of the program itself\footnote{Though this may not be the case--presumably you would need to be assured of the same inputs, or have the ability to check state up in until the moment of the execution of the game.}. But the implementation of the game itself offers further opportunities for a type of ``non-cooperation''. For instance, implementing a naive "Program Equilibrium" program in the game Matching Pennies would mean that the column program exploits the matching behavior of the row program. Some commitment to the game form itself appears to be needed here.

\end{document}